\documentclass[letterpaper, 10 pt, conference]{ieeeconf}

\IEEEoverridecommandlockouts                              %
\usepackage[T1]{fontenc}
\usepackage[utf8]{inputenc}

\usepackage{amsmath} %
\usepackage{amssymb} %

\usepackage{graphicx}
\usepackage{cite}

\usepackage{booktabs}

\usepackage{url}

\usepackage{hyperref}

\usepackage{xspace}

\newif\ifanonymous
\anonymousfalse

\title{\LARGE \bf EventGeM: Global-to-Local Feature Matching for\\Event-Based Visual Place Recognition}

\ifanonymous
  \author{Anonymous Author(s)%
  }
\else
  \author{Adam D. Hines$^{*}$ \quad Gokul B. Nair \quad Nicol\'{a}s Marticorena \quad Michael Milford \quad Tobias Fischer%
  \thanks{The authors are with the QUT Centre for Robotics, School of Electrical
  Engineering and Robotics, Queensland University of Technology, Brisbane, QLD 4000,
  Australia. $^{*}$Corresponding author: \texttt{adam.hines@qut.edu.au}}%
  \thanks{This work received funding from an ARC Laureate Fellowship FL210100156 to MM and an ARC Discovery Early Career Researcher Award DE240100149 to TF. The authors acknowledge continued support from the Queensland University of Technology (QUT) through the Centre for Robotics.}
  }
\fi

\begin{document}
\bstctlcite{IEEEexample:BSTcontrol}
\maketitle
\thispagestyle{empty}
\pagestyle{empty}

\begin{abstract}
Event cameras are rapidly rising in popularity for robotic and computer vision tasks because their sparse activation delivers energy-efficient, high-dynamic-range, and fast sensing. Event cameras have been used in robotic navigation and localization tasks where positioning must occur in real time with sufficient accuracy. However, current event-based localization methods suffer from poor spatial understanding and are not viewpoint tolerant. In this paper, we address the problem of viewpoint-robust place recognition directly from event streams. We present EventGeM, a global-to-local feature fusion pipeline for event-based visual place recognition that combines whole-image feature detection to shortlist top candidates for 2D homography-based re-ranking with random sample consensus (RANSAC). We also contribute a regional generalized mean pooling (GeM) layer that learns to return the most relevant spatial features using per-row exponents to pool event streams into a compact global descriptor, trained on the NYC-Event-VPR dataset. These contributions overcome shortfalls in currently available event-based localization methods that fail to recognize similar places with large changes in viewpoint. To evaluate viewpoint-robust localization, we contribute a new event-based dataset that includes repeated traverses with a severe lateral shift. EventGeM improves absolute Recall@1 by 7 to 43 percentage points over the strongest baseline in each experiment. We also deploy EventGeM on a robotic platform, demonstrating real-time performance of our hierarchical pipeline. The code for EventGeM is available at \url{https://github.com/AdamDHines/Event-GeM}.
\end{abstract}

\section{Introduction}
Visual place recognition (VPR) is a core component for robot localization and navigation, where incoming query images are matched against a reference database of images with known poses~\cite{Schubert2024}. State-of-the-art VPR systems use conventional frame-based images of places for feature extraction~\cite{Zhang2021, Masone2021}. Recently, there has been a growing interest in the use of event cameras, also known as dynamic vision sensors, to perform VPR due to their low-power, low-latency operation with high temporal resolution~\cite{Fischer2020, Lee2021, Fischer2022, Kong2022, Lee2023, Hines2025-2, Ramanathan2025}. 
In this work, we build an event-based VPR pipeline that carries lessons and implementations across from conventional systems~\cite{Masone2021}, and we show improved VPR performance relative to existing baseline methods~\cite{Hines2025-2, Lee2021, Fischer2022}.  

\begin{figure}[t]
  \centering
   \includegraphics[width=\columnwidth]{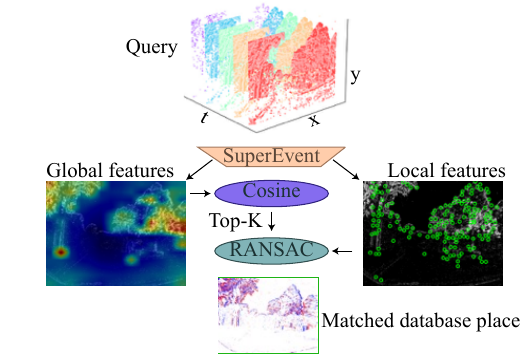}
   \caption{Schematic overview of EventGeM. A query event stream is converted to a multi-channel time surface (MCTS) and passed through the SuperEvent~\cite{Burkhardt2025} backbone. Global features are pooled by our multi-exponent generalized mean (GeM) layer~\cite{Radenovic2018} and ranked by cosine similarity to produce a top-$k$ shortlist. Local features from the SuperEvent keypoint detector head then re-rank the shortlist by 2D homography estimation with RANSAC, returning the matched database place.}
   \label{fig:eventgem}
   \vspace{-15pt}
\end{figure}

Whilst conventional frame-based VPR techniques benefit from a wide variety of pre-trained deep-learning computer vision models, such as from ResNet~\cite{He2015}, visual geometry group (VGG) networks~\cite{Simonyan2015}, or DINOv2~\cite{Oquab2024}, event-based data is not within the training domain of such vision models. This is because asynchronous event streams produce temporally rich information that is spatially sparse on microsecond timescales~\cite{Gallego2022}, which is fundamentally different from frame-based images with absolute pixel intensities at any given time. Simple representations such as binary or histogram counts of pixel-wise event activity over a fixed time window are among the most common ways of generating frames from asynchronous event camera streams for computer vision applications~\cite{Yang2023, Yang2024, Gallego2022, Hines2025}. Pre-trained models for event streams remain scarce, which limits the number of systems available to fine-tune for robotic localization. 

Here, we introduce a new event-based VPR pipeline called EventGeM (Fig.~\ref{fig:eventgem}) that leverages SuperEvent~\cite{Burkhardt2025}, a recently published event-native vision transformer trained for simultaneous localization and mapping. We repurpose the frozen SuperEvent backbone, a hierarchical attention network that interleaves local window and dilated grid attention, to produce global place descriptors, and we combine these with the native keypoint detector head to form a two-stage localization pipeline. To build compact global descriptors, we pool the SuperEvent feature map with a novel generalized mean (GeM)~\cite{Radenovic2018} layer that learns a separate pooling exponent per spatial region, retaining coarse spatial arrangement rather than collapsing the whole feature map into a single unstructured descriptor. This global stage produces an accurate top-$k$ shortlist, which we then re-rank by estimating a 2D homography over SuperEvent keypoint descriptors. EventGeM achieves the strongest VPR performance among the compared baselines~\cite{Lee2021, Fischer2022, Hines2025-2} across multiple benchmark datasets, including a newly contributed dataset, called Gardens-Point-Event, that features large deliberate viewpoint shifts, in contrast to prior datasets that focus on fixed on-the-rails traverses~\cite{Fischer2020, Carmichael2025}. Finally, we demonstrate real-time capabilities of our pipeline on a robotic platform.

Our contributions are as follows:
\begin{enumerate}
    \item We present EventGeM, an event-based VPR method that combines regional GeM pooling with learned exponents, a residual head trained for VPR, and 2D homography re-ranking of feature keypoints for end-to-end matching.
    \item We showcase real-time-capable localization of our proposed system, including re-ranking, with an on-robot demonstration using an event camera. 
    \item We contribute an event-camera dataset with repeated traverses under a severe lateral shift, and use it to demonstrate viewpoint robustness during localization.
\end{enumerate}

The code for EventGeM is avaiable at~\url{https://github.com/AdamDHines/Event-GeM}. The Gardens-Point-Event dataset is available for evaluation with instructions and full GPS annotation at~\url{https://huggingface.co/datasets/AdamHines/Gardens-Point-Event}.

\section{Related Work}
\label{sec:relatedworks}
In this section, we review conventional and state-of-the-art VPR systems in Sect.~\ref{subsec:vpr} and event-based localization and VPR methods in Sect.~\ref{subsec:eventvpr}.

\subsection{Visual place recognition}
\label{subsec:vpr}

VPR is an image retrieval task used in robotic navigation to match incoming query images to a reference database image with a known pose~\cite{Schubert2024}. SALAD~\cite{Izquierdo2024} fine-tuned the DINOv2 vision foundation model for optimal feature detection and introduced a new optimal transport aggregation technique for global descriptors, extended in MegaLoc with a rigorous training regime~\cite{Berton2025}. MixVPR~\cite{Ali2023} introduced a novel feature aggregation method based on a multi-layer perceptron (MLP) from multiple feature maps, with a focus on improving per-query runtimes. Like these methods, EventGeM aggregates a backbone feature map into a single compact descriptor, but it uses the event-native SuperEvent transformer~\cite{Burkhardt2025} as its backbone and aggregates with generalized mean pooling~\cite{Radenovic2018} under a learned per-region exponent rather than an MLP (Sect.~\ref{subsec:globalfeat}).

CosPlace~\cite{Berton2022} redesigned VPR as an image classification problem over position and heading, which circumvented slow training times often incurred for contrastive loss methods. As in CosPlace, EventGeM uses a batch-wise contrastive loss with positives defined by distance and heading, but EventGeM freezes the backbone and trains only the pooling exponents and a residual head (Sect.~\ref{subsec:training}). AP-GeM~\cite{Revaud2019} developed an average precision loss function with a GeM pooling layer using advances in listwise loss functions and histogram binning approximations. Learnable pooling exponents are already part of GeM~\cite{Radenovic2018} and AP-GeM~\cite{Revaud2019}, and pooling a feature map over spatial regions follows R-MAC~\cite{Tolias2016}. EventGeM adds one exponent per grid row, learned on an event backbone. 

Finally, EventGeM's global-to-local structure follows the frame-based practice of re-ranking a global shortlist by local feature matching and geometric verification~\cite{Masone2021, Schubert2024}, as in Patch-NetVLAD~\cite{Hausler2021} and R2Former~\cite{Zhu2023}; this stage is absent from existing event-based VPR methods, and EventGeM draws its keypoints from the same backbone as the global descriptor, so re-ranking needs no second network (Sect.~\ref{subsec:keypointrank}).

\subsection{Event-based localization}
\label{subsec:eventvpr}
Lee and Kim~\cite{Lee2021} introduced EventVLAD, one of the first deep-learned event-based VPR methods, which denoises reconstructed edge images before extracting NetVLAD features rather than operating on raw event frames. Sparse-Event-VPR explored the amount of pixel-wise information needed to perform effective VPR, showing that VPR can be performed with as few as 25 pixels~\cite{Fischer2022}. LENS deployed the first event-based VPR pipeline onto neuromorphic hardware for direct, asynchronous inferencing with off-chip readouts~\cite{Hines2025-2}. Recently, Flash performed sub-millisecond VPR using active pixel selection~\cite{Ramanathan2025}.

Alternatively, image reconstruction of RGB images from event streams~\cite{Rebecq19} can be used in conjunction with state-of-the-art vision transformers and VPR methods for accurate localization, with the additional cost of computationally expensive image generation. Event-LAB~\cite{Hines2025} showed that reconstructed RGB images outperform raw event frames by up to 50 percentage points, because the reconstructions admit conventional VPR techniques such as CosPlace~\cite{Berton2022}, EigenPlaces~\cite{Berton2023}, and MixVPR~\cite{Ali2023}. Joseph~\emph{et al.}~\cite{Joseph2025} showed that histogram-based localization can be enhanced by ensembling multiple techniques, including image-reconstruction-based methods. All of these methods retrieve in a single global step, without geometric verification, and are evaluated on on-the-rails traverses; EventGeM adds keypoint re-ranking that reuses the global backbone, and we evaluate it under a deliberate lateral shift (Sect.~\ref{subsec:viewvar}).

\section{Method}
\label{sec:methods}

We obtain spatial feature embeddings for initial place matches (Sect.~\ref{subsec:globalfeat}), pool them with a multi-patch GeM layer (Sect.~\ref{subsec:spatialpool}), re-rank the top-$k$ matches using SuperEvent keypoints (Sect.~\ref{subsec:keypointrank}), and train the model on NYC-Event-VPR~\cite{Pan2025} (Sect.~\ref{subsec:training}). 

\subsection{Spatial feature embeddings}
\label{subsec:globalfeat}
An event stream is represented as a sequence of events $e = (x, y, t, p)$, 
with pixel coordinates $(x,y)$, timestamp $t$, and polarity $p$. We partition the event stream into fixed temporal windows of duration \(\Delta t\) and convert each window into a multi-channel time surface (MCTS) representation~\cite{Burkhardt2025}, using the relative temporal decay scales \(\alpha_m \in \{0.01,0.03,0.1,0.3,1\}\) as implemented in~\cite{Burkhardt2025}, which give absolute time constants \(\tau_m = \alpha_m \Delta t\). The MCTS is therefore:

    \begin{align}
        \mathbb{M}_{x,y,p,m}
        =
        \exp\left(
            -\frac{
                t_{\mathrm{ref}} - t_{\mathrm{last}}(x,y,p)
            }{\tau_m}
        \right),
        \label{eq:mcts}
    \end{align}
    where $t_{\mathrm{ref}}$ is the reference time for a window slice, $m \in \{1,\dots,N_\tau\}$ indexes the $N_\tau$ time constants, and $t_{\mathrm{last}}(x,y,p)$ is the timestamp of the most recent event at location $(x,y)$ with polarity $p$. Locations for which no event is observed within the window are assigned a zero response. The resulting representation is $\mathbb{M} \in \mathbb{R}^{H \times W \times 2 \times N_\tau}$, where $H$ and $W$ are the height and width of the MCTS representation.

The MCTS is then passed through the SuperEvent $\mathbf{SE}$ vision-transformer backbone~\cite{Burkhardt2025} to obtain a spatial feature map $\mathcal{X}$:

    \begin{align}
        \mathcal{X}
        =
        \mathbf{SE}(\mathbb{M}),
        \qquad
        \mathcal{X}
        \in
        \mathbb{R}^{H_f \times W_f \times C},
        \label{eq:featuremap}
    \end{align}
where $H_f$ and $W_f$ are the spatial dimensions of the feature map and $C$ is its number of channels~\cite{Burkhardt2025}.

\subsection{Spatial feature pooling}
\label{subsec:spatialpool}

\begin{figure}[t]
      \centering
       \includegraphics[width=\columnwidth]{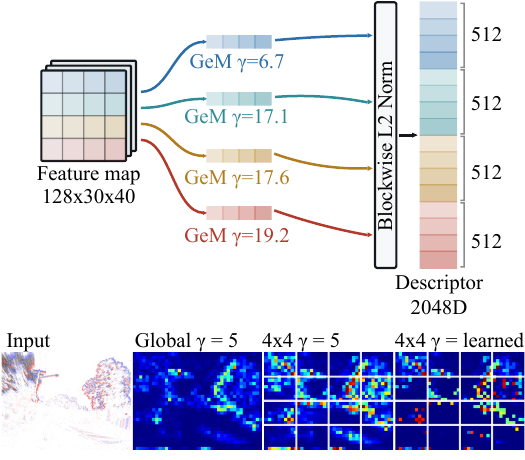}
       \caption{Multi-patch GeM pooling. (Top) The $128\times30\times40$ feature map is split into a $4\times4$ grid; each row is pooled with its own learned exponent $\gamma_r$ (values from Sect.~\ref{subsec:hyperparam}), giving four 512-dimensional blocks that are $\ell_2$-normalized and concatenated. (Bottom) Pooled activations for one event frame under a global $\gamma=5$, a $4\times4$ grid at $\gamma=5$, and a $4\times4$ grid with learned per-row exponents. Warmer colors denote stronger activation.}
       \label{fig:features}
       \vspace{-5pt}
    \end{figure}

To convert $\mathcal{X}$ into a compact place descriptor, we apply a generalized mean (GeM) pooling~\cite{Radenovic2018} layer over sixteen regions of the feature map (Fig.~\ref{fig:features}). The feature map is divided into a $4\times4$ grid indexed by row $r$ and column $v$, giving sixteen regions $\Omega_{r,v}$. For region $\Omega_{r,v}$ and exponent $\gamma_r$, the pooled response in channel $c$ is:

    \begin{align}
        g_{r,v,c}
        =
        \left[
            \frac{1}{|\Omega_{r,v}|}
            \sum_{(i,j)\in\Omega_{r,v}}
            {\mathcal{X}}_{i,j,c}^{\,\gamma_r}
        \right]^{1/\gamma_r},
        \label{eq:gem}
    \end{align}
where $c \in \{1,\dots,C\}$ indexes the feature channel. The feature map $\mathcal{X}$ is clamped at zero before pooling, because fractional exponents require non-negative inputs. We write $\mathbf g_{r,v} \in \mathbb{R}^{C}$ for the vector stacking $g_{r,v,c}$ over all channels.

We allow the pooling of features to follow the vertical structure of the scene, rather than pooling the complete feature map with a single global exponent. We achieve this by constructing the descriptor using \emph{multi-patch regional pooling} (Fig.~\ref{fig:features}). Regions in the same grid row share one exponent, so four exponents $\gamma_1,\ldots,\gamma_4$ are learned in total. This retains the coarse vertical structure that global pooling discards, and performs better than single exponents per patch (Sect.~\ref{subsec:ablation}).

Rather than fixing the exponents by hand, we treat them as learnable parameters optimized jointly with the retrieval objective (Sect.~\ref{subsec:training}). A softplus parameterization constrains each exponent to $\gamma_r \geq 1$, keeping the pooling a valid generalized mean, and we cap $\gamma_r \leq 20$ so that the pooling cannot collapse to a hard maximum during training. The learned values and the effect of this cap are reported in Sect.~\ref{subsec:hyperparam}.

Each pooled vector is independently $\ell_2$-normalized:
    \begin{align}
        \widehat{\mathbf g}_{r,v}
        =
        \frac{
            \mathbf g_{r,v}
        }{
            \left\|\mathbf g_{r,v}\right\|_2
        }.
        \label{eq:gem_norm}
    \end{align}

The sixteen regional descriptors are then concatenated:

    \begin{align}
        \mathbf h
        =
        \underset{r,v=1,\ldots,4}{\operatorname{concat}}
        \left(
            \widehat{\mathbf g}_{r,v}
        \right).
        \label{eq:multigem_concat}
    \end{align}

Finally, the complete descriptor is re-normalized:

    \begin{align}
        \mathbf f
        =
        \frac{\mathbf h}{\|\mathbf h\|_2},
        \qquad
        \mathbf f \in \mathbb{R}^{D}.
        \label{eq:multigem}
    \end{align}

Because the descriptor stacks sixteen regional $C$-dimensional vectors, its dimensionality is $D=16C$, which for the SuperEvent feature map gives $D=2{,}048$. The pooled descriptor is then refined by a learned residual head to give the final descriptor $\hat{\mathbf f}$, defined in Sect.~\ref{subsec:training}.

Given the final descriptors, we retrieve places by the cosine similarity $s_{q,i} = \hat{\mathbf f}_q^{\top} \hat{\mathbf f}_i$ between query $q$ and reference $i$, keeping the $k$ references with the highest $s_{q,i}$ as the shortlist $\mathcal{S}_k(q)$.

\subsection{Keypoint re-ranking}
\label{subsec:keypointrank}
For each query $q$ and each reference $i$ in its shortlist $\mathcal{S}_k(q)$, keypoints are detected using the pre-trained keypoint detector head of SuperEvent $\mathbf{SE}$ on the same MCTS place representation~\cite{Burkhardt2025}:

    \begin{align}
        \mathcal{K}^{q}, \mathbf{D}^{q} &= \mathbf{SE}(\mathbb{M}^{q}), \\
        \mathcal{K}^{i}, \mathbf{D}^{i} &= \mathbf{SE}(\mathbb{M}^{i}),
    \label{eq:se_kp_desc}
    \end{align}
    where $\mathcal{K}^{q}$ and $\mathcal{K}^{i}$ are the detected 2D keypoint locations in the query and the reference, respectively, and $\mathbf{D}^{q}$ and $\mathbf{D}^{i}$ are the corresponding local descriptors. From these, we compute a set of tentative descriptor matches $\mathcal{M}_{q,i}$ using a mutual nearest-neighbor (MNN) test, which keeps a pair only when each descriptor is the other's nearest neighbor,

    \begin{align}
        \mathcal{M}_{q,i}
        =
        \mathrm{MNN}
        \left(
            \mathbf{D}^{q},
            \mathbf{D}^{i}
        \right),
        \label{eq:nnr}
    \end{align}
    so that each match in $\mathcal{M}_{q,i}$ associates a descriptor from $\mathbf{D}^{q}$ with its corresponding descriptor in $\mathbf{D}^{i}$.

We estimate a homography $\mathbf{H}_{q,i}$ from these correspondences using RANSAC, where the set of geometrically verified inliers $\mathcal{I}_{q,i}$ is:
    \begin{align}
    \mathcal{I}_{q,i}
    =
    \left\{
    (\mathbf{p}^{q},\mathbf{p}^{i}) \in \mathcal{M}_{q,i}
    \ \middle|\
    \left\|
    \mathbf{p}^{i} - \pi\!\left(\mathbf{H}_{q,i}\,\tilde{\mathbf{p}}^{q}\right)
    \right\|_2
    < \epsilon
    \right\},
    \label{eq:ransac}
    \end{align}
    where $\tilde{\mathbf{p}}=[x,y,1]^\top$ denotes homogeneous coordinates, $\pi(\cdot)$ converts homogeneous coordinates back to Euclidean coordinates, and $\epsilon$ is the RANSAC re-projection threshold. A homography is exact only for a planar scene or a pure rotation, so we do not treat $\mathbf{H}_{q,i}$ as a metric estimate of viewpoint change; we use only $|\mathcal{I}_{q,i}|$, the number of matches admitting one consistent warp, as a geometric-consistency check.

Each shortlisted reference is then scored by the number of geometrically verified inliers it supports:
    \begin{align}
        s'_{q,i}
        =
        \left|\mathcal{I}_{q,i}\right|,
        \qquad
        i \in \mathcal{S}_k(q).
        \label{eq:rerank_score}
    \end{align}
    Because $\left|\mathcal{I}_{q,i}\right|$ is an integer, shortlisted references frequently tie; we break ties by the global similarity $s_{q,i}$, preserving the global ordering where geometric verification cannot separate two candidates. The final prediction for query $q$ is the shortlisted reference with the highest re-ranked score,
    \begin{align}
        \hat{\imath}(q)
        =
        \operatorname*{arg\,max}_{\,i \in \mathcal{S}_k(q)}
        \; s'_{q,i},
        \label{eq:kp_rerank}
    \end{align}
    and Recall@K is computed from the shortlist re-ordered by decreasing $s'_{q,i}$.

\subsection{Model training}
\label{subsec:training}
The regional pooling exponents $\{\gamma_r\}$ (\eqref{eq:gem}) and the residual head $\mathbf{MLP}$ were trained with a supervised contrastive learning rule on the NYC-Event-VPR dataset~\cite{Pan2025}, with the SuperEvent backbone frozen throughout. Each place $a$ carries a GPS fix and a heading. We write $d_{ab}$ for the great-circle distance between places $a$ and $b$, $\Delta\theta_{ab}$ for the circular heading difference, and $\theta_{\max}$ for the maximum permitted heading difference. A positive pair is formed by two places drawn from different recordings that lie within a radius $r_p$ of each other and agree in heading to within $\theta_{\max}$:

    \begin{align}
        \mathcal{P}(a)
        =
        \left\{\,
            b \;\middle|\;
            d_{ab} < r_p,\;
            c_b \neq c_a,\;
            \Delta\theta_{ab} < \theta_{\max}
        \,\right\},
        \label{eq:positives}
    \end{align}
    where $c_a$ indicates the recording from which place $a$ was derived. Negatives are places lying outside an exclusion radius $r_n$, so that $\mathcal{N}(a) = \left\{ b \mid d_{ab} \ge r_n \right\}$.

The contrastive loss for anchor $a$ is then defined as:

    \begin{align}
        \ell_a
        =
        \log \!\!\sum_{b' \in \mathcal{A}(a)}\!\!
            \exp\!\left(
            \frac{\hat{\mathbf f}_a^{\top} \hat{\mathbf f}_{b'}}{T}
            \right)
            \;-\;
            \frac{1}{|\mathcal{P}(a)|}
            \sum_{b \in \mathcal{P}(a)}
                \frac{\hat{\mathbf f}_a^{\top} \hat{\mathbf f}_b}{T},
        \label{eq:loss}
    \end{align}
    where $b$ indexes the positives $\mathcal{P}(a)$ of \eqref{eq:positives} and $b'$ indexes every candidate in $\mathcal{A}(a) = \mathcal{P}(a) \cup \mathcal{N}(a)$, the full set of positives and negatives available within the batch. The temperature $T$ scales the descriptor similarities.

The exponents and the head are optimized in two stages. First, the exponents $\{\gamma_r\}$ are learned without the head, with the loss applied directly to the pooled descriptor $\mathbf f$, so that they are fitted to the pooling itself rather than compensated for by the head. Each exponent is parameterized as $\gamma_r = 1 + \operatorname{softplus}(\rho_r)$, with $\rho_r$ the unconstrained parameter that is actually optimized. The exponents are then frozen and $\mathbf{MLP}$ is trained on the resulting descriptors. The head is a residual mapping $\hat{\mathbf f} = (\mathbf f + \mathbf{MLP}(\mathbf f)) / \|\mathbf f + \mathbf{MLP}(\mathbf f)\|_2$, with a zero-initialized final layer, so that training begins exactly at the raw pooled descriptor $\mathbf f$.

\section{Experimental Setup}
\label{sec:implementation}

In this section, we detail the hyperparameters for model training and inference (Sect.~\ref{subsec:hyperparam}), the baseline methods and datasets used for evaluation (Sect.~\ref{subsec:basemeth}), the evaluation metrics (Sect.~\ref{subsec:evalmet}), and finally the compute platforms used (Sect.~\ref{subsec:sysenviron}).

\subsection{Hyperparameters}
\label{subsec:hyperparam}
For model training, we used the contrastive loss of \eqref{eq:loss} and an AdamW optimizer with a learning rate of $3\times10^{-4}$ and a weight decay factor of $10^{-4}$. The $\mathbf{MLP}$ head was a $2048\times1568\times2048$ multi-layer perceptron with a dropout rate of 0.1. The temperature was set to $T=0.05$. Both training stages used a batch size of 1,024 over 500 epochs. We set $r_p = 25$\,m for positive pair selection and $r_n \ge 75$\,m for negative selection. We set a value for $\theta_{max}$ of 45${\text{o}}$.

The learned exponents increase monotonically down the grid, from $\gamma_1 \approx 6.7$ to $\gamma_4 \approx 19.2$. Three of the four (17.1, 17.6, and 19.2) sit within 15\% of the cap $\gamma_r \leq 20$ of Sect.~\ref{subsec:spatialpool}, so the cap binds: the lower three rows operate close to max pooling while only the top row retains a soft mean.

For the RANSAC re-ranking, we computed feature matches over the top $k=50$ shortlisted places, with a re-projection threshold $\epsilon=5.0$\,px.

\begin{table*}[!t]
  \footnotesize
  \caption{Recall at $K$ (R@K) on Brisbane-Event-VPR~\cite{Fischer2020}, with Sunset2 as the reference. \textbf{Bold} is best, \underline{underline} second best, per column. EventGeM (global only) is our descriptor before re-ranking. Bracketed values are the number of places.}
  \label{tab:brisbane_event}
  \begin{tabular*}{\textwidth}{@{\extracolsep{\fill}}l*{9}{c}@{}}
    \toprule
    & \multicolumn{3}{c}{Sunset1 (14{,}481)} & \multicolumn{3}{c}{Morning (13{,}456)} & \multicolumn{3}{c}{Daytime (14{,}321)} \\
    \cmidrule(lr){2-4}\cmidrule(lr){5-7}\cmidrule(lr){8-10}
    Method (Ref: Sunset2, 12{,}828) & R@1 & R@5 & R@10 & R@1 & R@5 & R@10 & R@1 & R@5 & R@10 \\
    \midrule
    LENS~\cite{Hines2025-2} & 0.09 & 0.16 & 0.20 & 0.05 & 0.10 & 0.14 & 0.04 & 0.08 & 0.13\\
    Sparse-Event-VPR~\cite{Fischer2022} & 0.05 & 0.08 & 0.13 & 0.02 & 0.04 & 0.05 & 0.05 & 0.08 & 0.09 \\
    EventVLAD~\cite{Lee2021} & 0.43 & 0.57 & 0.64 & 0.09 & 0.17 & 0.22 & 0.08 & 0.16 & 0.22 \\
    \midrule
    LENS~\cite{Hines2025-2}+Re-rank & 0.19 & 0.22 & 0.24 & 0.12 & 0.15 & 0.18 & 0.09 & 0.12 & 0.16 \\
    Sparse-Event-VPR~\cite{Fischer2022}+Re-rank & 0.11 & 0.12 & 0.15 & 0.03 & 0.04 & 0.05 & 0.05 & 0.09 & 0.10 \\
    EventVLAD~\cite{Lee2021}+Re-rank & 0.68 & 0.72 & 0.74 & 0.21 & 0.25 & 0.28 & 0.19 & 0.24 & 0.28 \\
    \midrule
    E2VID+AP-GeM~\cite{Rebecq19,Revaud2019} & 0.59 & 0.71 & 0.76 & 0.24 & 0.32 & 0.36 & 0.13 & 0.22 & 0.27\\
    ECDPT+GeM~\cite{Yang2023} & 0.77 & 0.86 & 0.90 & 0.29 & 0.48 & 0.58 & 0.19 & 0.36 & 0.45 \\
    SuperEvent~\cite{Burkhardt2025} & 0.86 & 0.89 & 0.90 & 0.53 & 0.67 & 0.73 & 0.37 & 0.51 & 0.59 \\
    \midrule
    EventGeM (global only) & \underline{0.91} & \underline{0.93} & \underline{0.94} & \underline{0.61} & \underline{0.74} & \underline{0.80} & \underline{0.58} & \underline{0.70} & \underline{0.76} \\
    EventGeM (ours) & \textbf{0.93} & \textbf{0.95} & \textbf{0.95} & \textbf{0.87} & \textbf{0.88} & \textbf{0.89} & \textbf{0.80} & \textbf{0.83} & \textbf{0.85} \\
    \bottomrule
  \end{tabular*}
\end{table*}

\subsection{Baseline methods and datasets}
\label{subsec:basemeth}
We used Event-LAB~\cite{Hines2025} to run the event-based baselines LENS~\cite{Hines2025-2}, Sparse-Event-VPR~\cite{Fischer2022}, and EventVLAD~\cite{Lee2021}. We additionally compare against a second event-native vision transformer (ViT) backbone, Event Camera Data Pre-training (ECDPT)~\cite{Yang2023}, with a GeM pooling layer (ECDPT+GeM). To isolate the value of the global stage, we also compare against SuperEvent~\cite{Burkhardt2025} with no global shortlist, matching keypoint descriptors exhaustively with a brute-force matcher from OpenCV. Finally, we compare against image reconstructions from E2VID~\cite{Rebecq19} paired with the conventional VPR method AP-GeM~\cite{Revaud2019}, which also uses a GeM pooling layer for image retrieval on a pre-trained ResNet101-AP-GeM backbone. 

We evaluate our method on the Brisbane-Event-VPR~\cite{Fischer2020} and the NSAVP~\cite{Carmichael2025} datasets. For each dataset, we performed all evaluations on multiple query traverses under different conditions with a single reference, following prior evaluation protocols in the literature~\cite{Berton2022, Hines2025}. All reference and query pairs are listed in the relevant results tables. A prediction counts as correct when it falls within a ground-truth tolerance of the query pose. We set this tolerance to 70\,m for the outdoor benchmark datasets~\cite{Fischer2020, Carmichael2025}, 25\,m for the Gardens-Point-Event dataset, and 3\,m for the indoor on-robot demonstration. 

For the Gardens-Point-Event viewpoint-variant dataset, we captured two walking traverses of an outdoor parkland and heritage-building precinct, separated by a lateral offset of approximately 5\,m. An iniVation DAVIS346 and an iPhone~16 acting as a GPS logger for ground-truth annotation were both mounted to a camera gimbal, and the two clocks were synchronized against an Ubuntu 24.04 laptop at the start of collection. Each traverse covered roughly 800\,m over 8\,min. The dataset will be released upon publication.

All MCTS representations for the main experiments were generated with a time window $\Delta t$ of 50\,ms. All event streams were resized to the input resolution of the SuperEvent backbone~\cite{Burkhardt2025} of $240\times320$ $H$ and $W$, respectively.

\subsection{Evaluation metrics}
\label{subsec:evalmet}
Our main evaluation metric is Recall@K, which measures whether any of the top $K$ matches predicted by a VPR system falls within the ground-truth tolerance. Recall $R$ is defined as:

    \begin{align}
        R = \frac{\mathrm{TP}}{\mathrm{GTP}},
        \label{eq:recall}
    \end{align}
    where $\mathrm{TP}$ is the number of true positive matches and $\mathrm{GTP}$ is the total number of possible ground-truth positive matches. We report Recall@K for $K \in \{1, 5, 10\}$ (R@1, R@5, R@10). We additionally report per-query throughput in Hz, measured end to end over the generation of the event representation and the execution of the respective VPR algorithm. 

\subsection{System environment}
\label{subsec:sysenviron}
EventGeM is implemented in Python and packaged with Pixi~\cite{Fischer2025}. Event streams, offline and online, were processed using EventCV~\cite{eventcv} with hot pixel filtering enabled. Model training and experiments were performed on an Ubuntu 24.04 desktop computer with an NVIDIA RTX~2080 graphics processing unit (GPU) with 8\,GB of memory. The on-robot demonstration used a 64\,GB Jetson Orin AGX Developer Kit running JetPack 6.2. 

\section{Results}
\label{sec:results}

Here, we evaluate the localization performance for EventGeM across benchmark event-based VPR datasets (Sect.~\ref{subsec:recall}), demonstrate viewpoint robustness on the contributed Gardens-Point-Event dataset with a deliberate lateral shift (Sect.~\ref{subsec:viewvar}), compare runtime with accuracy for all baseline methods (Sect.~\ref{subsec:runtime}), deploy EventGeM live on a mobile robotic platform (Sect.~\ref{subsec:onlinedemo}), and finally report the ablation studies (Sect.~\ref{subsec:ablation}).

\subsection{Recall performance of EventGeM}
\label{subsec:recall}

\begin{table}[t]
  \centering
  \footnotesize
  \setlength{\tabcolsep}{2pt}
  \caption{Recall at $K$ (R@K) on NSAVP~\cite{Carmichael2025}, with R0-FA0 as the reference. R0-FS0 is a daytime and R0-FN0 a night-time query. \textbf{Bold} is best, \underline{underline} second best, per column. EventGeM (global only) is our descriptor before re-ranking. Bracketed values are the number of places.}
  \label{tab:nsavp}
  \begin{tabular*}{\columnwidth}{@{\extracolsep{\fill}}p{0.42\columnwidth}*{6}{c}}
    \toprule
    & \multicolumn{3}{c}{R0-FS0 (18{,}218)} & \multicolumn{3}{c}{R0-FN0 (17{,}793)} \\
    \cmidrule(lr){2-4}\cmidrule(lr){5-7}
    Method (Ref: R0-FA0, 20{,}957) & R@1 & R@5 & R@10 & R@1 & R@5 & R@10 \\
    \midrule
    LENS~\cite{Hines2025-2}                     & 0.17 & 0.24 & 0.30 & 0.03 & 0.08 & 0.11 \\
    Sparse-Event-VPR~\cite{Fischer2022}         & 0.14 & 0.24 & 0.29 & 0.02 & 0.08 & 0.11 \\
    EventVLAD~\cite{Lee2021}                    & 0.20 & 0.28 & 0.33 & 0.05 & 0.11 & 0.14 \\
    \midrule
    LENS~\cite{Hines2025-2}+Re-rank             & 0.27 & 0.32 & 0.35 & 0.04 & 0.08 & 0.12 \\
    Sparse-Event-VPR~\cite{Fischer2022}+Re-rank & 0.19 & 0.27 & 0.31 & 0.04 & 0.08 & 0.11 \\
    EventVLAD~\cite{Lee2021}+Re-rank            & 0.34 & 0.38 & 0.40 & 0.05 & 0.10 & 0.13 \\
    \midrule
    E2VID+AP-GeM~\cite{Rebecq19,Revaud2019}     & 0.51 & 0.60 & 0.64 & 0.06 & 0.10 & 0.12 \\
    ECDPT+GeM~\cite{Yang2023}                   & 0.36 & 0.50 & 0.57 & 0.05 & 0.11 & 0.15 \\
    SuperEvent~\cite{Burkhardt2025}             & 0.34 & 0.46 & 0.51 & 0.05 & 0.12 & 0.17 \\
    \midrule
    EventGeM (global only)                      & \underline{0.75} & \underline{0.81} & \underline{0.83} & \underline{0.09} & \underline{0.16} & \underline{0.21} \\
    EventGeM (ours)                             & \textbf{0.83} & \textbf{0.85} & \textbf{0.86} & \textbf{0.15} & \textbf{0.21}  & \textbf{0.25} \\
    \bottomrule
  \end{tabular*}
  \vspace{-0.1cm}
\end{table}

To test our first claim, that EventGeM retrieves places more accurately than existing event-based methods under appearance change, Table~\ref{tab:brisbane_event} reports Recall@K on Brisbane-Event-VPR~\cite{Fischer2020}. For fairness we report the baselines both with and without re-ranking by SuperEvent keypoints and RANSAC over their own top-$k$ matches~\cite{Burkhardt2025}. EventGeM improves absolute R@1 over EventVLAD, the strongest event-native VPR baseline on this dataset, by 50 to 78 percentage points, and over SuperEvent, the strongest baseline overall, by 7 to 43 percentage points. The rows marked \emph{global only} give our descriptor before re-ranking, so the contribution of each stage can be read directly from the tables (Sect.~\ref{subsec:ablation}). LENS~\cite{Hines2025-2} and Sparse-Event-VPR~\cite{Fischer2022} have both been reported to perform poorly at short accumulation windows and to work best over longer time periods~\cite{Hines2025}, which is consistent with their low recall at the 50\,ms window used here.

On the NSAVP dataset~\cite{Carmichael2025}, EventGeM improves absolute R@1 over EventVLAD~\cite{Lee2021} by up to 63 percentage points and over E2VID+AP-GeM~\cite{Revaud2019,Rebecq19}, the strongest baseline on this dataset, by up to 32 percentage points, as summarized in Table~\ref{tab:nsavp}. The night-time pairing R0-FA0 versus R0-FN0 is the exception: EventGeM reaches only 15\% R@1 and every baseline falls between 2\% and 6\%. At night the event stream becomes sparse and noise-dominated, no longer resembling the daytime NYC-Event-VPR data the pooling and head were trained on. Adapting the camera biases to the lighting condition~\cite{Nair2024} is one remedy, and night-time operation remains a limitation shared by all compared methods.

\subsection{Viewpoint variant evaluation}
\label{subsec:viewvar}
\begin{figure}[!t]
      \centering
      \includegraphics[width=\columnwidth]{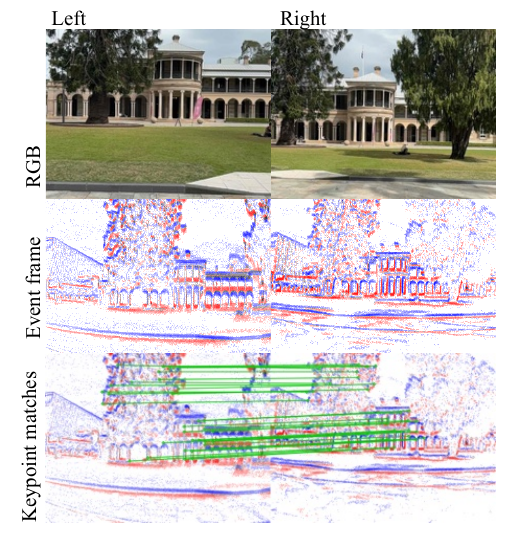}
       \caption{Example places from the left and right traverses of the Gardens-Point-Event dataset, separated by a lateral offset of approximately 5\,m. Top: RGB frames. Middle: event frames, with red and blue denoting positive and negative polarity. Bottom: green lines are mutual nearest-neighbor keypoint matches from SuperEvent~\cite{Burkhardt2025}.}
       \label{fig:gardensevent}
    \end{figure}

Existing event-based methods such as LENS~\cite{Hines2025-2} and Sparse-Event-VPR~\cite{Fischer2022} compare places pixel-wise, implicitly assuming that a repeat traverse reproduces the reference viewpoint. To test our third claim, that EventGeM tolerates viewpoint change, we collected a dataset by walking a repeated traverse with a deliberate lateral shift -- we call this Gardens-Point-Event (Fig.~\ref{fig:gardensevent}). Table~\ref{tab:qutevent} summarizes the recall performance of EventGeM and the baseline methods.

Recall drops for every method relative to the \emph{on-the-rails} benchmarks~\cite{Fischer2020, Carmichael2025}, confirming that the lateral shift makes the task harder. EventGeM still outperformed all baselines, reaching 55\% R@1 against 18\% for EventVLAD. The global descriptor alone reaches 37\%, so re-ranking contributes 18 percentage points, its third-largest gain across the six reference--query pairs. 

\begin{table}[t]
  \centering
  \footnotesize
  \setlength{\tabcolsep}{2pt}
  \caption{Recall at $K$ (R@K) on the Gardens-Point-Event dataset, Left as reference and Right as query. \textbf{Bold} is best, \underline{underline} second best, per column. EventGeM (global only) is our descriptor before re-ranking. Bracketed values are the number of places. Only the event-native baselines are shown to compare differences in viewpoint tolerance across established event baselines. 
}
  \label{tab:qutevent}
  \begin{tabular*}{\columnwidth}{@{\extracolsep{\fill}}p{0.42\columnwidth}*{3}{c}}
    \toprule
    & \multicolumn{3}{c}{Right (9{,}759)} \\
    \cmidrule(lr){2-4}
    Method (Ref: Left, 10{,}271) & R@1 & R@5 & R@10 \\
    \midrule
    LENS~\cite{Hines2025-2}                     & 0.09 & 0.14 & 0.21 \\
    Sparse-Event-VPR~\cite{Fischer2022}         & 0.06 & 0.14 & 0.21 \\
    EventVLAD~\cite{Lee2021}                    & 0.18 & 0.40 & 0.53 \\
    \midrule
    EventGeM (global only)                      & \underline{0.37} & \underline{0.58} & \underline{0.67} \\
    EventGeM (ours)                             & \textbf{0.55} & \textbf{0.70} & \textbf{0.75} \\
    \bottomrule
  \end{tabular*}
\end{table}

\subsection{Runtime performance}
\label{subsec:runtime}

    \begin{figure}[!t]
      \centering
       \includegraphics[width=\columnwidth]{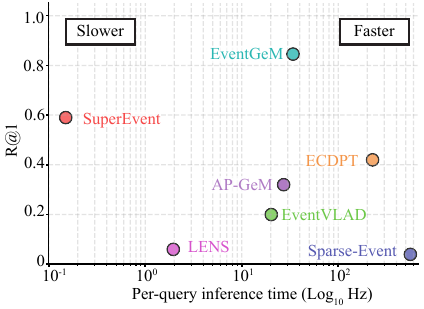}
       \vspace*{-6pt}
       \caption{Per-query throughput against average Recall@1 (R@1) on Brisbane-Event-VPR~\cite{Fischer2020} for EventGeM and the baselines~\cite{Hines2025-2,Fischer2022,Lee2021,Rebecq19,Revaud2019,Yang2023,Burkhardt2025}; methods further right are faster. Each point averages over the Sunset1, Morning, and Daytime queries with Sunset2 as the reference.}
       \label{fig:runtime}
       \vspace*{-6pt}
    \end{figure}
    
    \begin{figure*}[!t]
      \centering
       \includegraphics[width=0.8\textwidth]{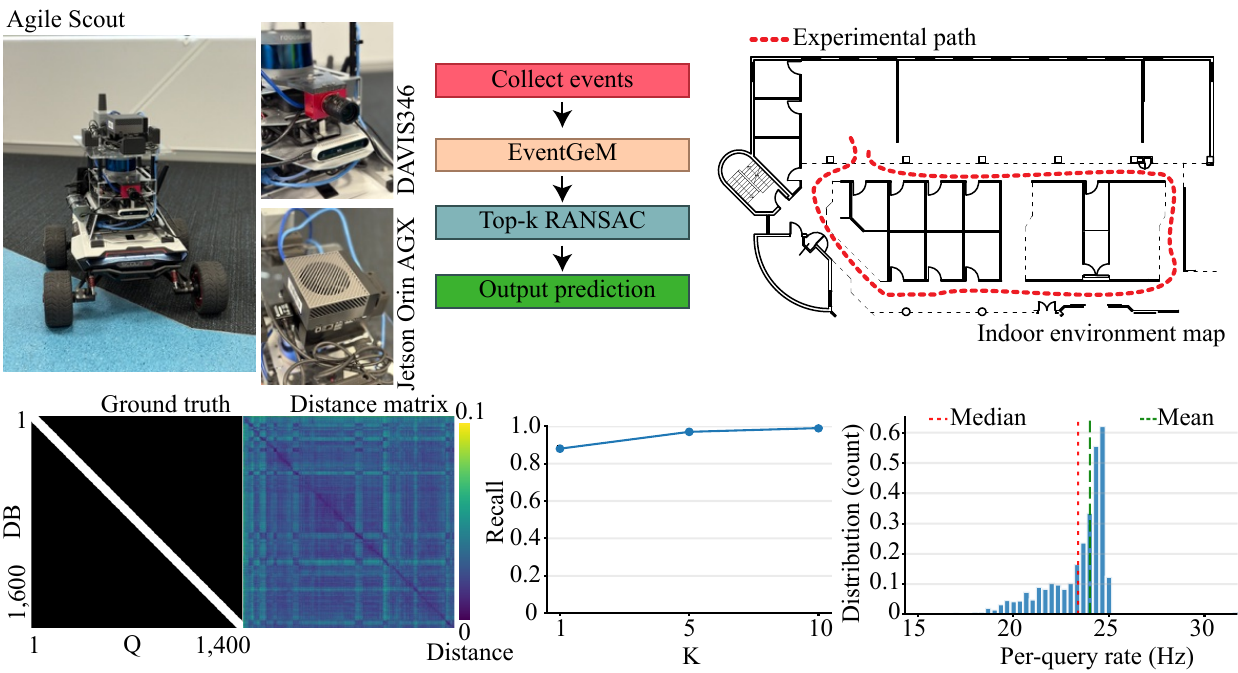}
       \small
       \caption{Online deployment of EventGeM on a robotic platform. Top, left to right: the AgileX Scout robot; its iniVation DAVIS346 dynamic vision sensor (DVS) and Jetson Orin AGX; the on-board pipeline; and the floor plan, with the teleoperated path in red, traversed once for the database and once as the query. Bottom, left to right: ground-truth correspondences between 1,400 queries (Q) and 1,600 database places (DB); the descriptor distance matrix, darker being closer; Recall@K; and per-query rate distribution, with median and mean marked.}
       \label{fig:robot}
    \end{figure*}
We next compared the per-query runtime of each method, measured end to end from streaming individual events out of the recording to producing a match (Fig.~\ref{fig:runtime}). EventGeM achieved real-time inference at 33.97\,Hz on the RTX~2080 desktop described in Sect.~\ref{subsec:sysenviron}. SuperEvent~\cite{Burkhardt2025} ran at only 0.15\,Hz, because it matches keypoint descriptors exhaustively against all 12,828 references. This highlights the cost that a global shortlist followed by top-$k$ re-ranking avoids. 

EventVLAD~\cite{Lee2021} ran at an average of 20.21\,Hz, and EventGeM raised the average R@1 over the three query traverses from 0.20 to 0.87 at a comparable throughput. LENS~\cite{Hines2025-2} and Sparse-Event-VPR~\cite{Fischer2022} achieved similar absolute recall to each other but at very different rates of 1.95\,Hz and 558\,Hz, respectively. E2VID+AP-GeM~\cite{Rebecq19, Revaud2019} ran at 27.17\,Hz, which includes the conversion from events to RGB frames. 

EventGeM offers the best balance of recall and throughput: Sparse-Event-VPR is an order of magnitude faster but reaches 0.04 average R@1, while SuperEvent, the only baseline within 0.3 R@1 of EventGeM, is over two hundred times slower.

\subsection{Online demonstration}
\label{subsec:onlinedemo}

Two-stage VPR models are accurate, but the cost of RANSAC re-ranking makes real-time operation difficult. To test our second claim, that the full pipeline including re-ranking runs online on embedded hardware, we deployed EventGeM on a robotic platform with an iniVation DAVIS346 and a Jetson Orin AGX, exporting the model to the Open Neural Network Exchange (ONNX) format and compiling it on the Jetson (Fig.~\ref{fig:robot}). The robot was teleoperated twice around an indoor environment, giving 1,600 database places and 1,400 queries. Against a 3\,m tolerance we obtained 88\% R@1, with the descriptor distance matrix following the ground-truth correspondence, and throughput averaged 24\,Hz per query. EventGeM is therefore suitable for on-board localization with event cameras.

\subsection{Ablation studies}
\label{subsec:ablation}

    \begin{table}[!t]
      \centering
      \caption{Effect of the MCTS accumulation window $\Delta t$ on Recall at $K$ (R@K) for EventGeM on the Brisbane-Event-VPR~\cite{Fischer2020} dataset. The reference traverse was Sunset2 and the query traverse was Morning.}
      \label{tab:timewindows}

      \footnotesize
    
      \begin{tabular}{lccc}
        \toprule
        $\Delta t$ (ms) & R@1 & R@5 & R@10 \\
        \midrule
        1   & 0.13 & 0.18 & 0.20 \\
        10  & 0.58 & 0.61 & 0.62 \\
        50  & 0.87 & 0.88 & 0.89 \\
        100 & 0.92 & 0.93 & 0.94 \\
        250 & 0.95 & 0.96 & 0.96 \\
        \bottomrule
      \end{tabular}
      \vspace*{-0.1cm}
    \end{table}

All ablations were performed on the Brisbane-Event-VPR dataset~\cite{Fischer2020} with Sunset2 as the reference and Morning as the query. We first examine the effect of the MCTS accumulation window $\Delta t$ on R@1, summarized in Table~\ref{tab:timewindows}. Increasing the time window of accumulation improves the R@1 of EventGeM, as has been observed for other event-based VPR methods~\cite{Hines2025}, at the cost of the available localization update frequency.

Table~\ref{tab:gemablate} shows that regional pooling gives the largest single gain over a global exponent, 0.59 to 0.81 R@1 at fixed $\gamma=5$, and that it does not continue past a $4\times4$ grid: the $8\times8$ variant is worse at four times the size. Learning one exponent per patch rather than per row reaches 0.82 R@1, within a point of the fixed-exponent grid, which is why we share along rows. The lower rows decompose the system: the learned-exponent descriptor alone reaches 0.55 R@1, the head adds 6 points before re-ranking and 8 after, and re-ranking adds 26 on top of the head.

Across the six reference--query pairs in Tables~\ref{tab:brisbane_event}, \ref{tab:nsavp}, and~\ref{tab:qutevent}, re-ranking adds between 2 points R@1 (Sunset1, where global retrieval is already near its ceiling) and 26 points (Morning), with the largest gains under appearance change and under the 5\,m lateral shift (18 points). Because re-ranking only permutes the top-50 shortlist, recall at the shortlist size is unchanged by construction; all of the gain comes from re-ordering candidates the global stage had already retrieved. 

    \begin{table}[t]
      \centering
      \caption{GeM pooling variants, evaluated on Brisbane-Event-VPR~\cite{Fischer2020} with Sunset2 as reference and Morning as query; all models trained on NYC-Event-VPR~\cite{Pan2025} per Sect.~\ref{subsec:training}. \textbf{Bold} is best, \underline{underline} second best, per column. R@K is Recall at $K$ and \emph{Dim.} the descriptor dimensionality. Rows marked \emph{global} are scored before re-ranking.}
      \label{tab:gemablate}
    
      \setlength{\tabcolsep}{4pt}
      \renewcommand{\arraystretch}{1.12}
      \footnotesize
    
      \begin{tabular}{@{}lcccc@{}}
        \toprule
        Condition & R@1 & R@5 & R@10 & Dim. \\
        \midrule
        $\gamma = 1$ & 0.32 & 0.36 & 0.39 & 128 \\
        $\gamma = 2$ & 0.53 & 0.57 & 0.58 & 128 \\
        $\gamma = 3$ & 0.59 & 0.63 & 0.65 & 128 \\
        $\gamma = 5$ & 0.59 & 0.64 & 0.66 & 128 \\
        \midrule
        $\gamma = 5$ + $2\times2$ patch & 0.78 & 0.80 & 0.81 & 512 \\
        $\gamma = 5$ + $4\times4$ patch & 0.81 & \underline{0.83} & 0.83 & 2048 \\
        $\gamma = 5$ + $8\times8$ patch & 0.77 & 0.78 & 0.79 & 8192 \\
        \midrule
        learned $4\times4$, per-patch $\gamma$ & \underline{0.82} & \underline{0.83} & \underline{0.84} & 2048 \\
        \midrule
        learned $4\times4$, no head, global & 0.55 & 0.66 & 0.71 & 2048 \\
        learned $4\times4$, no head, re-rank & 0.79 & 0.81 & 0.81 & 2048 \\
        learned $4\times4$ + head, global & 0.61 & 0.74 & 0.80 & 2048 \\
        learned $4\times4$ + head, re-rank (ours)\hspace{-2pt} & \textbf{0.87} & \textbf{0.88} & \textbf{0.89} & 2048 \\
        \bottomrule
      \end{tabular}
    \end{table}

\section{Conclusions}
\label{sec:conclusions}
We present EventGeM, an event-based VPR system that leverages pre-trained vision transformers and keypoint detectors for accurate localization. Across multiple datasets, EventGeM improved R@1 by 7 to 43 percentage points over the strongest baseline in each experiment, the exception being night-time NSAVP, where every method fails. We also contribute a dataset with repeated traverses under a severe lateral shift, on which EventGeM reaches 55\% R@1 against 18\% for the best baseline, and we deploy the full pipeline end to end on a robot at 24\,Hz.

Despite these encouraging results, there is further space for improvement. Night-time operation remains unsolved for every method we evaluated, the planar homography behind our re-ranking score only approximates general outdoor scenes, and the frozen backbone is never adapted to place recognition. Adapting the camera biases to the lighting condition~\cite{Nair2024}, learning the geometric score, and fine-tuning the backbone on a large-scale event VPR dataset are the clearest routes to further gains.

\footnotesize
\bibliography{references}

\end{document}